\pdfoutput=1
\documentclass{article}
\PassOptionsToPackage{table}{xcolor}
\usepackage{iclr2027_conference,times}

\usepackage{amsmath,amsfonts,bm}

\def\Secref#1{Section~\ref{#1}}

\def\eqref#1{equation~\ref{#1}}

\def\1{\bm{1}}

\DeclareMathAlphabet{\mathsfit}{\encodingdefault}{\sfdefault}{m}{sl}
\SetMathAlphabet{\mathsfit}{bold}{\encodingdefault}{\sfdefault}{bx}{n}

\usepackage{xcolor}
\usepackage{graphicx}
\usepackage{booktabs}
\usepackage{multirow}
\usepackage{makecell}
\usepackage{caption}
\usepackage{enumitem}
\usepackage{algorithm}
\usepackage{algpseudocode}
\usepackage{hyperref}
\usepackage{url}

\newcommand{\ours}{CoEx}
\newcommand{\base}{CoEx w/o code agent}
\newcommand{\baseshort}{w/o code agent}

\usepackage{listings}
\definecolor{promptbg}{HTML}{F7F8FA}
\definecolor{promptframe}{HTML}{D0D7DE}
\definecolor{prompttitle}{HTML}{24292F}
\definecolor{prompttitlebg}{HTML}{E8ECF0}

\lstdefinestyle{prompt}{
  basicstyle=\fontsize{7.5}{9.2}\selectfont\ttfamily,
  breaklines=true,
  breakatwhitespace=false,
  columns=fullflexible,
  keepspaces=true,
  escapeinside={(*@}{@*)},
  backgroundcolor=\color{promptbg},
  frame=single,
  framesep=6pt,
  framerule=0.5pt,
  rulecolor=\color{promptframe},
  xleftmargin=6.5pt,
  xrightmargin=6.5pt,
  aboveskip=6pt,
  belowskip=6pt,
}
\newcounter{prompt}
\renewcommand{\theprompt}{\arabic{prompt}}
\DeclareCaptionFormat{promptfmt}{\hspace*{-6.5pt}\makebox[\textwidth]{\hfil#1#2#3\hfil}}
\DeclareCaptionLabelFormat{promptlabel}{Prompt~\theprompt}
\lstnewenvironment{promptbox}[1][]{%
  \refstepcounter{prompt}%
  \lstset{style=prompt,
    captionpos=b,
    caption={[{#1}]{#1}},
    abovecaptionskip=4pt,
    belowcaptionskip=4pt,
    label={prompt:\theprompt},
  }%
  \captionsetup{type=figure,format=promptfmt,labelformat=promptlabel,labelsep=colon,font=small}%
}{}

\graphicspath{{figures/}}

\title{CoEnv: Code-Driven Exploration in\\ Real-to-Sim Digital Twins for\\ Multi-Arm Manipulation}

\author{Li Kang$^{1,2\star}$\quad Yutao Fan$^{2,3\star}$\quad Rui Li$^{2,3\star}$\quad Heng Zhou$^{2,4\star}$\quad Yiran Qin$^{5}$\\
\textbf{Zhemeng Zhang$^{1}$\quad Songtao Huang$^{2,6}$\quad Xiufeng Song$^{1}$\quad Zaibin Zhang$^{7}$}\\
\textbf{Bruno N.Y. Chen$^{8}$\quad Zhenfei Yin$^{9}$\quad Dongzhan Zhou$^{2\dagger}$\quad Wangmeng Zuo$^{3\dagger}$\quad Lei Bai$^{2\dagger}$}\\[4pt]
{\normalfont $^{1}$Shanghai Jiao Tong University\quad $^{2}$Shanghai AI Laboratory\quad $^{3}$Harbin Institute of Technology}\\
{\normalfont $^{4}$University of Science and Technology of China\quad $^{5}$CUHK-Shenzhen\quad $^{6}$Fudan University}\\
{\normalfont $^{7}$Dalian University of Technology\quad $^{8}$Carnegie Mellon University\quad $^{9}$University of Oxford}\\[4pt]
{\normalfont $^{\star}$Equal contribution\quad $^{\dagger}$Corresponding authors}
}

\iclrfinalcopy

\begin{document}

\maketitle
\lhead{Preprint}

\begin{abstract}
General-purpose coding agents can now write, debug, and refine entire robot programs, but only if they have somewhere to practice. Physical hardware is a poor practice ground: exploratory rollouts are slow to reset and, when several arms share a workspace, unsafe. Generic simulation benchmarks are safe but do not match the geometry, assets, or viewpoints of the workspace where the robot must actually work. We present \textbf{\ours{}} (\textbf{Co}de-driven \textbf{Ex}ploration), which lets the agent practice in an \emph{executable digital twin of its own deployment workspace}. The twin is built by \textbf{CoEnv}, our real-to-sim framework, which reconstructs the physical scene from multi-view RGBD into a physics simulation. Inside it, a centralized code agent explores at the \emph{program level}, writing a complete multi-arm program, running it in the twin, and refining it against grounded verification (privileged state checks, checkpoint predicates on risky steps, and adaptive viewpoints against inter-arm occlusion). Verified programs transfer to hardware through collision-checked deployment. On five real-world multi-arm tasks with up to three heterogeneous robots, \ours{} reaches 70\% average success, against 50\% for the same system without the code agent (where a VLM must choose every primitive step by step), outperforms two prior VLM-planning frameworks, and produces verified episodes at $3.8$--$11.7\times$ the ablated rate. Because every verified rollout is also a demonstration, exploration leaves reusable experience behind: a $\pi_{0.5}$ policy fine-tuned purely on 200 agent-generated rollouts reaches 100\% and 70\% real-robot success on two tasks, with no human demonstrations.
\end{abstract}

\section{Introduction}
\label{sec:intro}

\begin{figure}[t!]
  \centering
  \includegraphics[width=\textwidth]{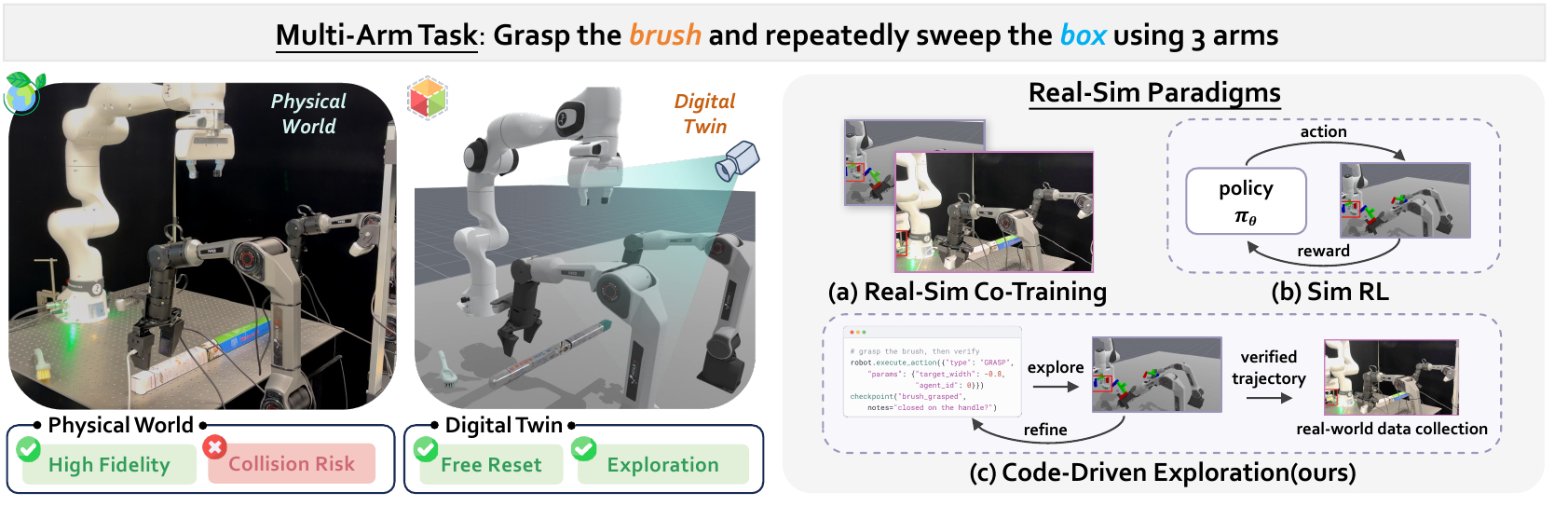}
  \vspace{-8pt}
  \caption{\textbf{Where does the simulator sit?} \textit{Left:} we reconstruct the multi-arm workcell into an executable digital twin. The real cell is faithful, but every exploratory attempt risks a collision; the twin resets freely and can be explored at no cost. \textit{Right:} three ways of using that twin. \textbf{(a)}~co-training mixes simulated trajectories into the training set alongside a few real ones, and \textbf{(b)}~sim RL optimizes a policy $\pi_\theta$ against simulated rollouts and a reward. Both place the simulator upstream of the \emph{data}, so the policy learns from simulated experience and a sim-to-real gap has to be bridged. \textbf{(c)}~\ours{} places it upstream of the \emph{program}: given only a task description, the code agent writes a multi-arm program, explores and refines it in the twin until it verifies, and the verified trajectory is then executed on the real robot, so the data collected afterwards is real.}
  \label{fig:motivation}
  \vspace{-12pt}
\end{figure}

Visuomotor policies are bounded by the supply of demonstrations, and nowhere more severely than in multi-arm manipulation, where a single teleoperated demonstration can occupy several synchronized operators and recent systems still keep a human in the loop for every episode~\citep{lin2026hats,zhao2024roco}. The standard answer is simulation: one line multiplies a few human demonstrations into a large simulated corpus~\citep{mandlekar2023mimicgen,jiang2025dexmimicgen,qin2026comsim}, another co-trains on simulated and real data~\citep{maddukuri2025simreal}, a third reconstructs the scene and learns inside it~\citep{torne2024rialto,dai2024automated}. All of them place the simulator upstream of the \emph{data} (Fig.~\ref{fig:motivation}a,b), with two consequences: the policy learns from simulated experience, so a sim-to-real gap has to be bridged; and the behavior being simulated still originates from a human, as seed demonstrations or as a reward. The bottleneck is displaced rather than removed.

Coding agents suggest a different route. Vision-language and code-generation models decompose tasks, write control programs, inspect outcomes, and correct their own mistakes~\citep{liang2023code,zhou2025code,zawalski2024robotic,huang2024rekep}, and general-purpose coding agents now write, debug, and refine entire robot programs~\citep{tsui2026faea,elmaaroufi2026rho,lu2026aspire,xiao2026enpire}. An agent that can solve a multi-arm task on its own needs neither a demonstration nor a reward. But the edit--run--debug loop that makes such agents effective needs somewhere to run the program, see what went wrong, and try again, cheaply and often. Hardware is neither: an agent that retries spends its budget on resets, risks collisions between arms sharing a workspace, and cannot see past arms that occlude one another. Generic simulation benchmarks are safe but do not match the geometry, assets, or camera views of the deployment workspace, so programs debugged there do not transfer.

Our key idea is to give the agent an \emph{executable digital twin of its own workspace} (Fig.~\ref{fig:motivation}). We reconstruct the specific physical scene (objects, poses, robot placements, camera calibration) into a physics simulation (real-to-sim), in which a coding agent can run programs, render arbitrary viewpoints, query privileged state, and verify outcomes at zero physical risk. The twin plays the role of a world model in model-based learning~\citep{hafner2023dreamerv3,bruce2024genie}, an action-conditioned predictor in which behavior is tried before it is executed, but it is \emph{reconstructed rather than learned}: dynamics come from a physics engine and state is explicit. This is deliberate: debugging a program requires ground-truth state (poses, contacts, swept volumes) that tells the agent \emph{what} went wrong, which learned video world models do not expose. Because the twin is metrically aligned with the real workcell, verified programs transfer back to hardware through collision-checked deployment. The twin is therefore used to find a \emph{program}, never to produce training images: once a program is verified, re-running it on the real robot turns one round of exploration into a corpus of real demonstrations, collected without a teleoperator and with no gap left to bridge (Fig.~\ref{fig:motivation}c).

We instantiate this idea as \textbf{\ours{}}, code-driven exploration hosted by \textbf{CoEnv}, a real-to-sim framework that builds the twin and returns verified programs to hardware. \emph{(i)~Digital twin construction} converts multi-view RGBD observations into a simulator-ready scene via generated 3D assets, open-vocabulary segmentation with VLM disambiguation, 6-DoF pose fusion, and iterative camera calibration (\Secref{sec:environment-digitization}). \emph{(ii)~Code-driven exploration (\ours{})}: a centralized code agent (one model controlling several embodiments, not decentralized multi-agent autonomy) writes a complete multi-arm program, runs it in the twin, and refines it against grounded verification, supported by twin-side tools for inter-arm occlusion and risky preconditions; as a baseline we remove the code agent (\base{}), leaving a VLM to choose every primitive step by step on the same twin (\Secref{sec:cognitive-planning-with-embodied-grounding}). \emph{(iii)~Collision-aware deployment} admits trajectories to hardware only after swept-volume checks across all arm pairs (\Secref{sec:validated-sim-to-real-transfer}). \emph{(iv)~Experience reuse}: verified rollouts persist as exemplars for future planning and as supervision for visuomotor policies (\Secref{sec:distillation}).

We evaluate \ours{} on five real-world multi-arm tasks with heterogeneous embodiments. \ours{} reaches 70\% average real-world success (up to 90\%), ahead of the same system without the code agent (50\%) on all five tasks and of two prior VLM-planning frameworks, at $3.8\times$--$11.7\times$ the ablated system's verified-episode throughput; once a \ours{} program clears its first milestone it completes the task in 88\% of trials, against 57\% without the code agent; the twin's adaptive-camera and checkpoint tools prove individually indispensable ($1.7\times$ and $2.5\times$ over their ablations); and no arm--arm collision occurred in 100 deployed trajectories. As a closing study, a $\pi_{0.5}$ policy fine-tuned purely on 200 \ours{} rollouts reaches 100\% and 70\% real-robot success on two tasks with no human demonstrations, up from 80\% and 40\% with 50 rollouts.

Our main contributions are:
\begin{itemize}[leftmargin=1.4em,itemsep=2pt,topsep=2pt,parsep=0pt]
\item \textbf{A different place for the simulator}: it searches for a \emph{program} rather than generating data, so multi-arm demonstrations can be collected on the real robot with no teleoperator, no reward, and no sim-to-real gap in the training set.
\item \textbf{\ours{}}, code-driven exploration in which a code agent writes, runs, and refines complete multi-arm programs against grounded verification, supported by twin-side tools for inter-arm occlusion (adaptive camera control) and risky preconditions (checkpoint predicates), and compared against the same system without the code agent (\base{}).
\item \textbf{Real-world validation} on five multi-arm tasks with up to three heterogeneous robots via collision-checked deployment, with comparisons to prior frameworks, tool ablations, and failure analysis.
\item \textbf{Evidence that the resulting data is what matters}: policies fine-tuned on \ours{} rollouts beat those trained on human teleoperation and, at matched size, on rollouts recorded in the twin instead of on the robot.
\end{itemize}

\section{Related Work}
\label{sec:related-work}

\noindent\textbf{Agentic planning and coding agents for robots.}
Foundation models plan robot behavior through code generation~\citep{liang2023code,li2024robocoder}, execution monitoring~\citep{zhou2025code}, keypoint constraints~\citep{huang2024rekep}, chain-of-thought action reasoning~\citep{zawalski2024robotic,zhen20243d}, and simulation-grounded planning~\citep{liu2025simpact}. A fast-moving line hands the task to general-purpose \emph{coding agents}, which solve benchmark manipulation from privileged state~\citep{tsui2026faea}, search policy repositories from execution feedback~\citep{elmaaroufi2026rho}, compound validated fixes into skill libraries~\citep{lu2026aspire}, and run policy improvement on robot fleets~\citep{xiao2026enpire}. \ours{} belongs to this line, and practices in a twin of the \emph{actual deployment workspace} rather than in benchmark scenes, with verification built for multi-arm occlusion and contact. LLM coordination has also been extended to multiple robots through dialog or role assignment~\citep{zhao2024roco,singh2025malmm,qin2025robofactory,gong2024mindagent,palma2026bicicle}, largely on textual abstractions without geometric grounding~\citep{gerkey2004taxonomy,liu2022embodied}. \ours{} instead uses a deliberately \emph{centralized} planner over all arms; multi-arm manipulation is our testbed, not a claim of decentralized autonomy.

\noindent\textbf{Simulation, digital twins, and where the simulator sits.}
Because teleoperation does not scale, much work uses simulation to supply training data: multiplying a few demonstrations into thousands of simulated trajectories~\citep{mandlekar2023mimicgen,jiang2025dexmimicgen,qin2026comsim}, co-training a large simulated corpus with a small real dataset~\citep{maddukuri2025simreal}, or reconstructing the deployment scene, learning inside it, and transferring with domain randomization~\citep{torne2024rialto,dai2024automated,lou2026drex,zhu2025vr}. All place the simulator upstream of the \emph{data} (Fig.~\ref{fig:motivation}; Appendix~\ref{sec:supp-positioning} tabulates them), so a sim-to-real gap must be bridged and the simulated behavior still originates from a human, as seed demonstrations or as a reward. Both bind hardest with multiple arms, where one demonstration occupies several synchronized operators and recent systems keep a human in the loop for every episode~\citep{lin2026hats}. \ours{} places the simulator upstream of the \emph{program} instead: the agent finds and verifies a multi-arm program with neither demonstrations nor rewards, and that program then runs on the real robot, so the data it leaves behind is real. Hence our twin is \emph{reconstructed rather than learned}: world models predict observations from data~\citep{hafner2023dreamerv3,bruce2024genie,yang2024unisim} but expose no privileged state against which success can be checked, whereas workspace-aligned physics simulators~\citep{todorov2012mujoco,mittal2025isaac,tao2024maniskill3,mu2021maniskill,nasiriany2024robocasa,chen2023frsr,wan2025lodestar,chen2025robotwin} give the ground-truth poses, contacts, and swept volumes that tell the agent \emph{what} went wrong.

\noindent\textbf{Agent-generated robot data.}
Foundation models also generate robot training data by synthesizing tasks and scenes~\citep{wang2024robogen,wang2024gensim}, designing rewards~\citep{ma2024eureka,ma2024dreureka}, collecting autonomously at fleet scale~\citep{ahn2024autort}, and distilling LLM-guided rollouts into policies~\citep{ha2023scaling,duan2024manipulate}; language agents likewise improve by accumulating skills and reflections~\citep{wang2023voyager,shinn2023reflexion}, and robot foundation models by autonomous practice~\citep{ghasemipour2025self}. Demonstration quantity drives policy performance~\citep{lin2024data}, so such generation is increasingly attractive. In \ours{}, data is a by-product rather than the goal: the rollouts the agent verifies while solving tasks are themselves demonstrations.

\section{Method}
\label{sec:method}

\begin{figure}[!t]
  \centering
  \includegraphics[width=\textwidth]{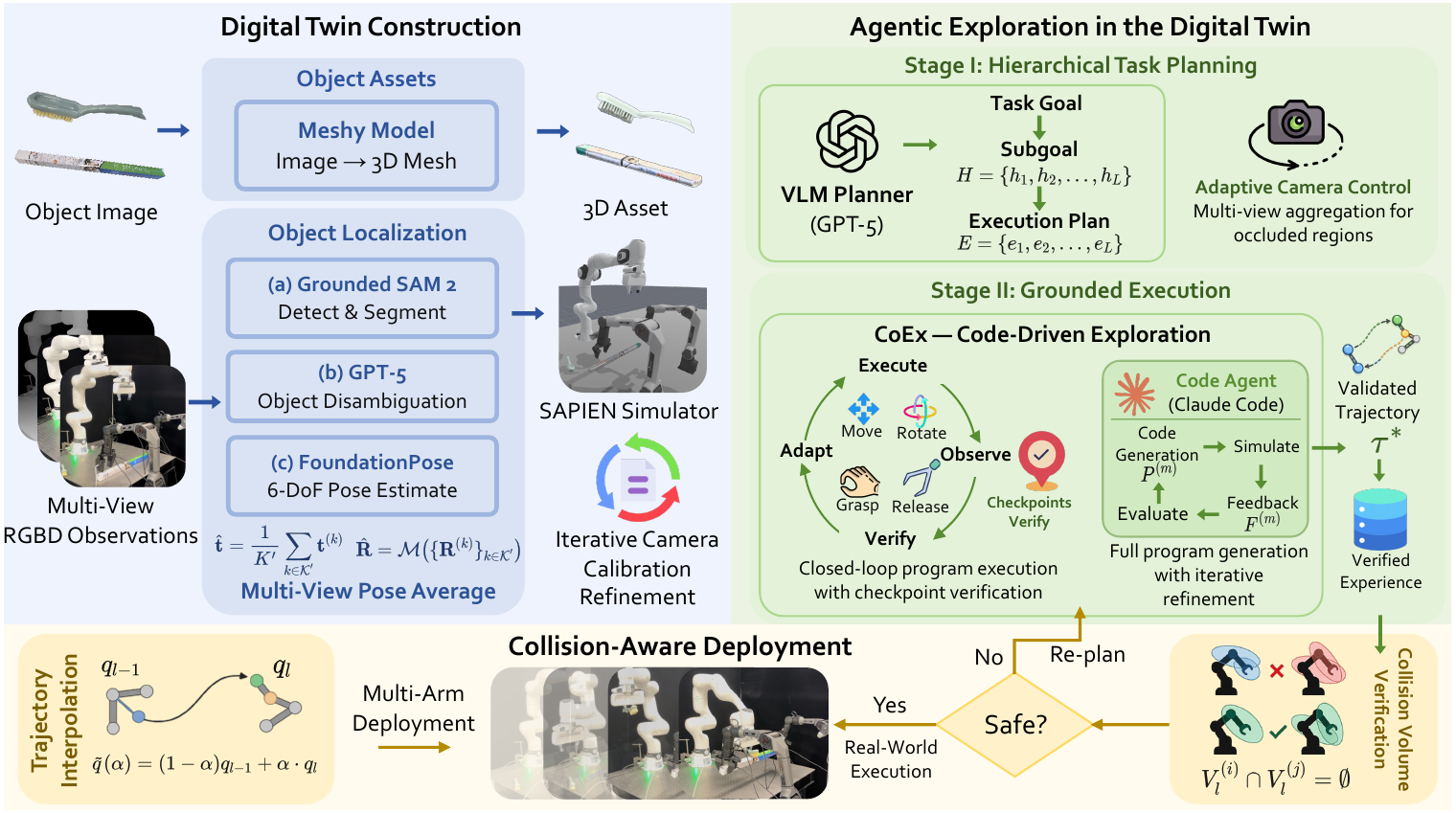}
  \caption{\textbf{The CoEnv framework.} \textbf{Top-left:} Digital twin construction converts multi-view RGBD into an executable simulation. \textbf{Top-right:} Agentic exploration: after hierarchical planning, \ours{} couples a program-level loop, in which a code agent generates, simulates, and refines a complete multi-arm program, with the step-level execute--observe--verify--adapt loop that running the program closes over primitives and checkpoints; validated trajectories are stored as verified experience $\mathcal{D}$. \textbf{Bottom:} Collision-aware deployment via interpolation with swept-volume verification.}
  \label{fig:pipeline}
  \vspace{-6pt}
\end{figure}

CoEnv couples a reconstructed digital twin with code-driven agentic exploration and collision-checked deployment; verified rollouts persist as reusable experience (Fig.~\ref{fig:pipeline}). We denote the twin $\mathcal{W}=(\mathcal{S},f,\textsc{Render})$, comprising the structured state space $\mathcal{S}$, the physics transition function $\mathbf{s}_{t+1}=f(\mathbf{s}_t,\mathbf{a}_t)$, and a rendering function over arbitrary camera poses.

\subsection{Digital Twin Construction from Real Scenes}
\label{sec:environment-digitization}

From \(K\) calibrated RGBD views \(\mathbf{o}_t\) we build a structured simulator state \(s_t = \Phi(\mathbf{o}_t)\). Simulation-ready meshes are generated from a few reference images per object and paired with convex collision proxies and per-category physical attributes; Grounded SAM2~\citep{ren2024grounded} segments the objects, GPT-5.6 Sol~\citep{openai2026gpt56} disambiguates visually similar instances, and FoundationPose~\citep{wen2024foundationpose} gives a 6-DoF pose per view, which we fuse across views by averaging translations and quaternions~\citep{markley2007averaging}. We implement \(\Phi\) in ManiSkill~\citep{tao2024maniskill3} on SAPIEN~\citep{xiang2020sapien}, whose contact resolution is stable at the millimeter scale, and refine camera extrinsics by rendering and comparing against the real capture until reprojection error stabilizes (Appendix~\ref{sec:supp-twin}). What matters downstream is not photorealism but that $\mathcal{W}$ is resettable, queryable, and metrically aligned with the workcell: the agent debugs against ground-truth state, and the training images come from the real cameras, never from $\mathcal{W}$.

\subsection{Code-Driven Exploration with Grounded Verification}
\label{sec:cognitive-planning-with-embodied-grounding}

Given state \(\mathbf{s}_t\), a centralized planner synthesizes a joint action \(\mathbf{a}_t=[a_t^{(1)},\dots,a_t^{(N)}]\) for \(N\) arms evolving under $f$. Given a goal \(\mathcal{G}\), exploration proceeds in two stages (Fig.~\ref{fig:pipeline}, top-right): \emph{hierarchical planning} decomposes \(\mathcal{G}\) into a structured plan, and \emph{grounded execution} realizes it through Code-driven Exploration (\ours{}), in which a program is written, run, and refined against verification. Both exploit affordances unavailable in the physical world: free resets, arbitrary rendered viewpoints, and privileged state queries.

\noindent\textbf{Stage I: Hierarchical task planning.}
A VLM planner (GPT-5.6 Sol) decomposes the goal into a two-level structure
$\mathcal{G}\!\to\!\mathcal{H}\!=\!\{h_l\}_{l=1}^{L}\!\to\!\mathcal{E}\!=\!\{e_l\}_{l=1}^{L}$,
where each \(h_l\) is a semantic sub-goal (e.g., ``pick up the red cube'') and each \(e_l=(i_l,\rho_l,\mathbf{p}_l)\) assigns arm \(i_l\) an action primitive \(\rho_l\!\in\!\{\textsc{Move},\textsc{Rotate},\textsc{Grasp},\textsc{Release}\}\) with parameters \(\mathbf{p}_l\), executed via inverse kinematics.

\begin{algorithm}[t]
    \caption{\ours{}: code-driven exploration in the digital twin}
    \label{alg:action-synthesis}
    \small
    \begin{algorithmic}[1]
    \Require Twin $\mathcal{W}$ with initial state \( \mathbf{s}_0 \), goal \( \mathcal{G} \), round budget \( M_{\max} \), experience set $\mathcal{D}$
    \Ensure Verified program \( \mathcal{P}^* \) and trajectory \( \tau^* \)
    \State \( \hat{o}_0 \gets \textsc{Aggregate}(\{\textsc{Render}(\mathbf{s}_0, c_j)\}_{j=1}^{J}) \) \Comment{adaptive multi-view observation}
    \State \( \mathcal{H} \gets \textsc{Decompose}(\mathcal{G}, \hat{o}_0) \); \( \mathcal{E} \gets \textsc{Assign}(\mathcal{H}, \mathbf{s}_0) \) \Comment{Stage I: hierarchical plan}
    \State \( \mathcal{F}^{(0)} \gets \textsc{Retrieve}(\mathcal{G}, \mathbf{s}_0, \mathcal{D}) \) \Comment{hints from past verified episodes}
    \For{\( m = 1, \dots, M_{\max} \)}
        \State \( \mathcal{P}^{(m)} \gets \textsc{CodeGen}(\mathbf{s}_0, \mathcal{E}, \mathcal{F}^{(m-1)}) \) \Comment{complete multi-arm program}
        \State \( \tau^{(m)} \gets \textsc{Execute}(\mathcal{P}^{(m)}, \mathcal{W}) \) \Comment{free reset; checkpoints snapshot state and views}
        \If{\( \textsc{Verified}(\tau^{(m)}, \mathcal{G}) \)} \( \mathcal{P}^* \gets \mathcal{P}^{(m)},\ \tau^* \gets \tau^{(m)} \); \textbf{break}
        \Else \ \( \mathcal{F}^{(m)} \gets \textsc{Analyze}(\tau^{(m)}, \text{logs}, \text{checkpoints}) \) \Comment{structured failure feedback}
        \EndIf
    \EndFor
    \State \( \mathcal{D} \gets \mathcal{D} \cup \{(\mathbf{s}_0, \mathcal{G}, \tau^*)\} \); re-execute \( \mathcal{P}^* \) from new initial states to add further verified rollouts
    \State Deploy \( \tau^* \) to hardware after swept-volume collision checks (\Secref{sec:validated-sim-to-real-transfer})
    \end{algorithmic}
\end{algorithm}

\begin{figure}[t]
\begin{lstlisting}[style=prompt,basicstyle=\fontsize{7.2}{8.8}\selectfont\ttfamily,frame=single]
# privileged state of the twin
robot.get_robot_state(robot_id)  -> {position, orientation, euler_rpy, gripper}
robot.get_object_state(name)     -> {position, orientation}
# primitives via robot.execute_action(a); agent_id: int or list (synchronized arms)
a = {"type": "MOVE",    "params": {delta_x, delta_y, delta_z, agent_id}}
a = {"type": "ROTATE",  "params": {delta_yaw, delta_pitch, delta_roll, agent_id}}
a = {"type": "GRASP",   "params": {target_width, agent_id}}
a = {"type": "RELEASE", "params": {agent_id}}
# grounded verification: snapshot state + multi-view renderings at a critical step
checkpoint(name, checkpoint_type, notes, suggested_views)
\end{lstlisting}
\vspace{-8pt}
\caption{\textbf{Interface exposed to \ours{} in the twin} (abridged from Appendix~\ref{sec:supp-code-prompts}). Programs read privileged state, issue delta-pose primitives to one or several arms at once, and place checkpoints whose snapshots drive failure analysis.}
\label{fig:api}
\vspace{-6pt}
\end{figure}

\noindent\textbf{Stage II: Grounded execution with \ours{}.}
We expose the primitive library as a code interface (Fig.~\ref{fig:api}) and perform \textbf{Co}de-driven \textbf{Ex}ploration (\ours{}), which couples the two loops drawn in Fig.~\ref{fig:pipeline} (Algorithm~\ref{alg:action-synthesis}). The \emph{outer, program-level loop} belongs to a code agent (Claude Code;~\citealp{anthropic2025claudecode}): it generates a complete program \(\mathcal{P}^{(m)}=\textsc{CodeGen}(\mathbf{s}_0,\mathcal{E},\mathcal{F}^{(m-1)})\) encoding the full multi-arm execution logic, simulates it in $\mathcal{W}$ to obtain \(\tau^{(m)}\), evaluates the outcome, and turns failures into structured feedback \(\mathcal{F}^{(m)}\) (collision locations, unachieved sub-goals) that conditions the next program, up to \(M_{\max}\) rounds. The \emph{inner, step-level loop} is what simulating a program means: the program \textit{executes} primitives on one or several arms, \textit{observes} privileged state and rendered views, \textit{verifies} outcomes and checkpoint predicates at critical junctures, and \textit{adapts} the following motions to what it observed; the logs and checkpoint snapshots it leaves behind are exactly what the outer loop evaluates. Because each round resets $\mathcal{W}$ for free, the agent can afford genuine trial-and-error: the twin converts exploration from a physical liability into cheap test-time search. And because a verified program is a reusable artifact, re-executing it from new initial states yields further verified rollouts without further deliberation, so exploration doubles as a data engine (\Secref{sec:exp-distill}).

\noindent\textbf{Why the program is the unit of exploration.}
The two loops differ in what one act of deliberation commits to. A step-level agent commits to a single primitive, observes, and deliberates again, so an episode costs one model call per step and the reasoning evaporates when the episode ends. A program-level agent commits to an entire multi-arm execution: deliberation is amortized over the whole task, a failure is diagnosed against the program that produced it rather than against an isolated step, and what survives the episode is source code. That artifact is why exploration compounds here. It can be re-executed from new initial configurations without further deliberation, retrieved as an exemplar when a related goal appears, and read by a person before it ever reaches hardware.

\noindent\textbf{Grounded verification tools.}
The twin exposes three tools to any agent acting in it. A verifier \(v_l = V(e_l,\mathbf{s}_t,\mathbf{s}_{t'})\in\{\textsc{Success},\textsc{Fail}\}\) checks each outcome against privileged state. \emph{Checkpoints} \(e_l^{\textsc{ckpt}}\) are non-action steps at critical junctures (e.g., gripper alignment before a grasp) that snapshot state and multi-view renderings and evaluate a predicate \(\phi(e_l^{\textsc{ckpt}},\mathbf{s}_t)\in\{0,1\}\). And because multi-arm workspaces are frequently occluded by the arms themselves, \emph{adaptive camera control} queries multiple poses \(\{c_j\}_{j=1}^{J}\) and aggregates renderings, $\hat{o}_t=\textsc{Aggregate}(\{\textsc{Render}(\mathbf{s}_t,c_j)\}_{j=1}^{J})$, to expose occluded regions.

\noindent\textbf{Ablated baseline (\base{}).}
To measure what the program-level loop buys, we remove the code agent: \base{} keeps the same twin, primitives, and tools but runs the inner loop alone, with a VLM (GPT-5.6 Sol) choosing every primitive step by step: it re-parameterizes \(e_l\) or inserts corrective steps when $v_l=\textsc{Fail}$, and proceeds past a checkpoint only when \(\phi=1\). This design reacts within an episode, but pays for deliberation at every primitive and leaves no reusable artifact.

\subsection{Collision-Aware Real-World Deployment}
\label{sec:validated-sim-to-real-transfer}

We bridge the kinematics gap between $\mathcal{W}$ and physical robots via \emph{trajectory interpolation} and \emph{collision volume verification}. For each arm \(i\), we record joint configurations \(\mathbf{q}_l^{(i)}\!\in\!\mathbb{R}^d\) at the completion of every primitive in simulation, and densify the trajectory by linear interpolation
$\tilde{\mathbf{q}}^{(i)}(\alpha)=(1-\alpha)\mathbf{q}_{l-1}^{(i)}+\alpha\,\mathbf{q}_{l}^{(i)}$,
discretized into \(S\) uniform steps for the real-world controller. Before each interpolated action executes, a forward-kinematics pass derives the swept collision volume \(\mathcal{V}_l^{(i)}\) for every arm from its link geometries along the interpolated path. An action is admitted only if \(\mathcal{V}_l^{(i)}\!\cap\!\mathcal{V}_l^{(j)}\!=\!\emptyset\) for all \(i\neq j\); otherwise it is discarded and re-planning is triggered from the current state. This pre-execution check keeps multi-arm operation collision-free without overly conservative motion constraints.

\subsection{Reusing Verified Experience}
\label{sec:distillation}

Exploration leaves a by-product: every verified rollout enters an experience set (``Verified Experience'' in Fig.~\ref{fig:pipeline})
\(\mathcal{D}=\{(\mathbf{s}_0^{(j)},\mathcal{G}^{(j)},\tau^{(j)})\}_{j=1}^{|\mathcal{D}|}\), which can be reused in two ways.
\emph{As exemplars:} the framework stores every verified episode, and when \ours{} faces goal $\mathcal{G}$ in state $\mathbf{s}_0$ it receives relevant past experience as in-context hints (strategies and grasp parameters that survived verification, and feedback from failed rounds). This mechanism is active in all our experiments; we do not ablate it separately.
\emph{As supervision:} each verified trajectory unrolls into pairs $\{(o_t,\mathbf{a}_t)\}_{t=1}^{T}$, where $o_t$ comprises the camera images and proprioception recorded while the verified program executes, and we fine-tune a visuomotor policy $\pi_\theta$ by behavior cloning on action chunks,
\begin{equation}
\mathcal{L}_{\textsc{BC}}(\theta) = \mathbb{E}_{(o_t,\,\mathbf{a}_{t:t+H})\sim \mathcal{D}}\bigl[\,\ell\bigl(\pi_\theta(o_t),\, \mathbf{a}_{t:t+H}\bigr)\bigr],
\label{eq:bc}
\end{equation}
where $\ell$ is the policy's native imitation loss and $H$ the chunk horizon. Unlike the agent, whose programs query privileged simulator state, $\pi_\theta$ acts from raw observations alone and replaces per-episode deliberation with a forward pass. We test this use of the data by fine-tuning $\pi_{0.5}$~\citep{pi2025pi05} in a small study (\Secref{sec:exp-distill}); re-verifying policy rollouts with the same tools to grow $\mathcal{D}$ further is a natural extension we leave to future work.

\section{Experiments}
\label{sec:experiments}

Our experiments answer four questions. \textbf{Q1:} Does code-driven exploration in the twin solve real multi-arm tasks, and how does it compare with step-wise deliberation (\Secref{sec:exp-real})? \textbf{Q2:} How does \ours{} compare with prior agentic planning frameworks (\Secref{sec:exp-baselines})? \textbf{Q3:} Which tools make exploration reliable, and where does it fail (\Secref{sec:exp-ablation})? \textbf{Q4:} Is the verified experience that exploration leaves behind reusable (\Secref{sec:exp-distill})?

\subsection{Setup}
\label{sec:experimental-setup}

\noindent\textbf{Hardware and tasks.}
We evaluate on five multi-arm manipulation tasks in two workcell configurations (Fig.~\ref{fig:tasks}; full milestone definitions in Appendix~\ref{sec:supp-tasks}). \textit{Two-arm} (Franka Research~3 $\times$2): \textbf{Cube Stacking}, \textbf{Ball Pickup} (bimanual lift), and \textbf{Transfer Cylinder} (handover). \textit{Three-arm} (Franka + AgileX Piper dual-arm): \textbf{Place Cucumber} (lid lifting + insertion) and \textbf{Brush Box} (coordinated sweeping). Robot bases are fixed; object poses vary across trials. Each workcell uses 2--3 calibrated RealSense D435i cameras, and scenes are reconstructed in ManiSkill (\Secref{sec:environment-digitization}).

\noindent\textbf{Agents and protocols.}
\ours{} is driven by Claude Code as the code agent, with a budget of $M_{\max}{=}10$ refinement rounds per task. The \base{} baseline shares the twin, primitives, and verification tools but lets GPT-5.6 Sol choose each primitive, so the two differ in backbone model as well as in loop structure. Real-world evaluations use 10 trials per agent per task. We report subtask milestones $S_i$ and task success rate (SR).

\subsection{Real-World Multi-Arm Evaluation (Q1)}
\label{sec:exp-real}

\begin{table}[t]
  \centering
  \caption{\textbf{Real-world multi-arm evaluation.} 10 trials per agent per task. $S_i$: $i$-th subtask milestone ($x$/10). SR: task success. Bold: higher SR per task.}
  \label{tab:main_results}
  \footnotesize
  \renewcommand{\arraystretch}{1.15}
  \setlength{\tabcolsep}{4pt}
  \begin{tabular*}{\textwidth}{@{\extracolsep{\fill}} l cccc cccc @{}}
    \toprule
    & \multicolumn{4}{c}{\textbf{\ours{} (ours)}} & \multicolumn{4}{c}{\textbf{\base{} (baseline)}} \\
    \cmidrule(lr){2-5} \cmidrule(lr){6-9}
    Task & $S_1$ & $S_2$ & $S_3$ & SR & $S_1$ & $S_2$ & $S_3$ & SR \\
    \midrule
    \rowcolor{gray!6}
    \multicolumn{9}{l}{\textit{Two-Arm (Franka $\times$ 2)}} \\
    Cube Stacking      & 10/10 & 9/10 & {--} & \textbf{9/10} & 7/10  & 6/10  & {--}  & 6/10 \\
    Ball Pickup        & 6/10  & 6/10 & {--} & \textbf{6/10} & 9/10  & 4/10  & {--}  & 4/10 \\
    Transfer Cylinder  & 8/10  & 6/10 & 6/10 & \textbf{6/10} & 9/10  & 4/10  & 4/10  & 4/10 \\
    \midrule
    \rowcolor{gray!6}
    \multicolumn{9}{l}{\textit{Three-Arm (Franka + AgileX Piper $\times$ 2)}} \\
    Place Cucumber     & 8/10  & 8/10 & 6/10 & \textbf{6/10} & 9/10  & 7/10  & 4/10  & 4/10 \\
    Brush Box          & 8/10  & 8/10 & 8/10 & \textbf{8/10} & 10/10 & 9/10  & 7/10  & 7/10 \\
    \midrule
    \rowcolor{gray!6}
    \textit{Average}   & {80\%} & {74\%} & {67\%} & \textbf{70\%} & {88\%} & {60\%} & {50\%} & {50\%} \\
    \bottomrule
  \end{tabular*}
  \vspace{-6pt}
\end{table}

\begin{figure}[t]
  \centering
  \includegraphics[width=\textwidth]{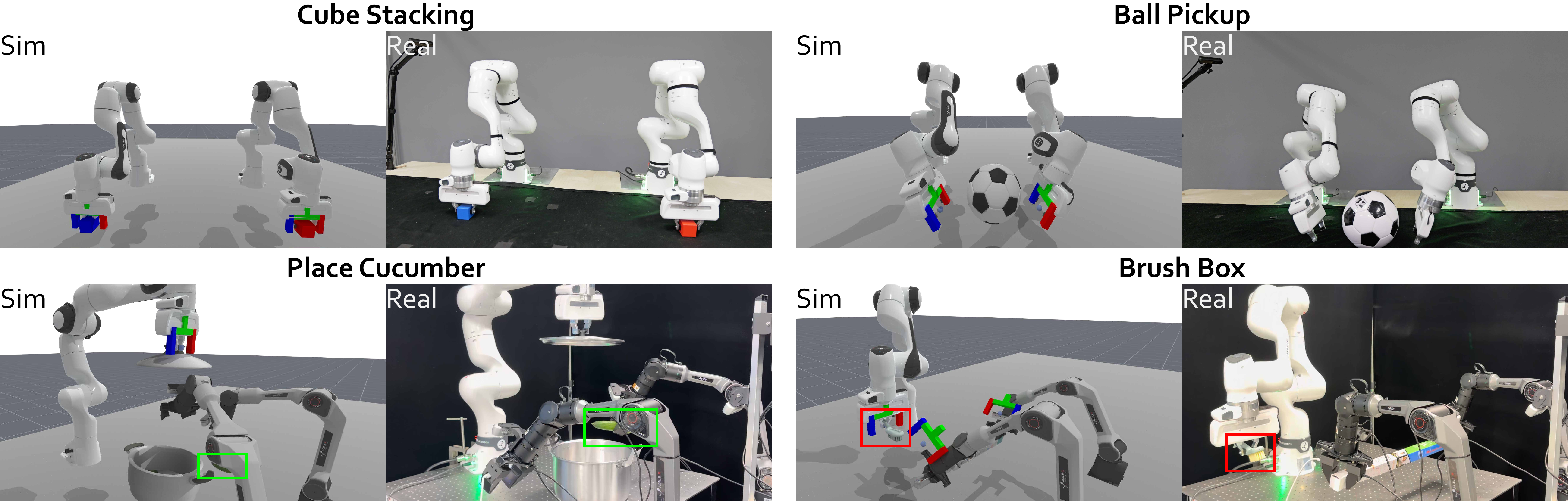}
  \caption{\textbf{Digital twin to real world.} Planned twin rollouts beside real executions on four tasks.}
  \label{fig:sim_real}
  \vspace{-8pt}
\end{figure}

Table~\ref{tab:main_results} reports real-robot results. \ours{} reaches 70\% average SR, against 50\% without the code agent, and is ahead on all five tasks, peaking at 90\% on \textit{Cube Stacking}; with 10 trials per cell, per-task gaps are best read as trends. The more telling pattern is \emph{where} each system loses trials. The step-wise ablation clears the first milestone more often (88\% vs.\ 80\%) but sheds successes at every later stage, finishing only 25 of the 44 trials that pass $S_1$ (57\%). \ours{} finishes 35 of 40 (88\%): every \textit{Brush Box} trial that grasps the brush completes the sweep (8/8/8), and the two three-stage tasks that involve a handover or a held lid keep 6 of 8. Two thirds of \ours{}'s failures (10 of 15) occur before the first milestone: on \textit{Ball Pickup}, all four failures happen at the initial bimanual contact, and every trial that establishes it completes the lift. A program that has been verified end to end in the twin encodes the whole multi-arm sequence coherently, so it either fails early, at the first grasp, or carries the task through, whereas per-step decisions let small errors compound along the horizon. Fig.~\ref{fig:sim_real} shows the correspondence between twin rollouts and physical execution, and no arm--arm collision occurred across all 100 deployed trajectories.

\begin{figure}[t]
  \centering
  \includegraphics[width=\textwidth]{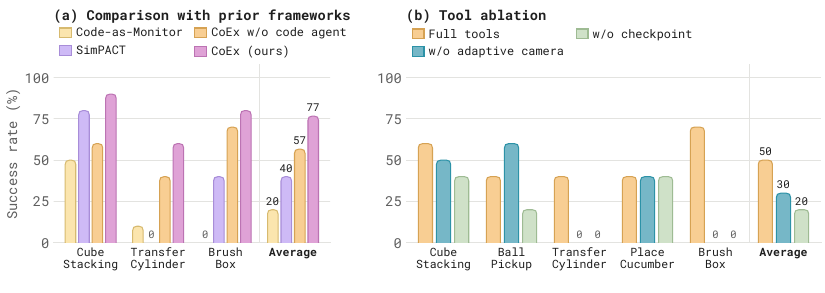}
  \vspace{-14pt}
  \caption{\textbf{Real-world success rates} (10 trials per task). \textbf{(a)}~\ours{} and our \base{} baseline against two prior VLM-planning frameworks on three representative tasks (easy two-arm, reactive two-arm handover, three-arm). \textbf{(b)}~Removing either twin-side tool from \base{}. Per-task counts are tabulated in Appendix~\ref{sec:supp-bars}.}
  \label{fig:bars}
  \vspace{-8pt}
\end{figure}

\subsection{Comparison with Prior Agentic Planning Frameworks (Q2)}
\label{sec:exp-baselines}

Fig.~\ref{fig:bars}(a) compares against two prior VLM-planning frameworks and our \base{} baseline on three representative tasks. Code-as-Monitor~\citep{zhou2025code} performs execution-time failure detection without sim-grounded planning; SimPACT~\citep{liu2025simpact} grounds planning in simulation but lacks an iterative verify--refine loop. \ours{} reaches 77\% on these tasks, against 57\% without the code agent and 20\% and 40\% for the prior frameworks. The gap is widest on the \textit{Transfer Cylinder} handover, where both prior frameworks collapse (1/10 and 0/10) while \ours{} completes 6/10. The two fail for complementary reasons: Code-as-Monitor detects a failure only once it has happened on hardware, and SIMPACT commits to the first plan its simulator accepts. \ours{} meets the same failures inside the twin, where retrying is free.

\subsection{What Makes Exploration Reliable, and Where It Fails (Q3)}
\label{sec:exp-ablation}

\noindent\textbf{Tool ablation.}
Fig.~\ref{fig:bars}(b) isolates the two twin-side tools on \base{}, where each tool's effect on individual decisions is directly observable. Removing \textbf{adaptive camera control} drops average SR from 50\% to 30\%, concentrated on occlusion-heavy tasks (\textit{Transfer Cylinder} and \textit{Brush Box} both fall to 0/10) where acting arms obstruct the contact region from any fixed viewpoint. Removing \textbf{checkpoint verification} is more damaging (50\%$\to$20\%): without precondition gating, early errors compound through long horizons. The two address complementary failure modes of exploration: \emph{observability} and \emph{precondition reliability}.

\noindent\textbf{Failure analysis.}
Failures fall into three recurring modes: residual sim-to-real pose offsets at grasp or placement, re-planning that stagnates on a failed strategy, and agent-specific limitations (drift over long step-wise sequences, and handovers that a single program must anticipate). The most direct remedies are a residual correction at contact and a hybrid that hands reactive sub-steps to step-wise control while the rest of the task stays at the program level; Appendix~\ref{sec:supp-failure} discusses representative cases.

\subsection{Reusing Verified Experience (Q4)}
\label{sec:exp-distill}

\noindent\textbf{Throughput.}
Table~\ref{tab:data-collection} reports verified episodes per agent session: \ours{} yields 17.5 and 9.5 on \textit{Cube Stacking} and \textit{Brush Box}, $11.7\times$ and $3.8\times$ the \base{} baseline (1.5 and 2.5), because one verified program amortizes reasoning across many rollouts, whereas a step-wise agent pays per primitive. Twin resets are instantaneous.

\begin{table}[t]
  \centering
  \caption{\textbf{Reusable experience.} \textit{Left:} verified episodes per agent session. \textit{Right:} real-robot SR (10 trials) of a $\pi_{0.5}$ policy fine-tuned on human teleoperated demonstrations, on those demonstrations co-trained with 200 rollouts of the same verified program recorded \emph{in the twin}, or on 50 and 200 rollouts of that program recorded \emph{on the robot}. Bold: best per row.}
  \label{tab:data-collection}
  \footnotesize
  \renewcommand{\arraystretch}{1.15}
  \setlength{\tabcolsep}{4pt}
  \begin{tabular*}{\textwidth}{@{\extracolsep{\fill}} l cc cccc @{}}
    \toprule
    & \multicolumn{2}{c}{\textbf{Verified episodes / session}} & \multicolumn{4}{c}{\textbf{$\pi_{0.5}$ policy SR by training data}} \\
    \cmidrule(lr){2-3} \cmidrule(lr){4-7}
    Task & \baseshort{} & \ours{} & 20 teleop. & \makecell{200 in twin\\+ 20 teleop.} & 50 \ours{} & 200 \ours{} \\
    \midrule
    Cube Stacking & 1.5 & \textbf{17.5} & 4/10 & 3/10 & 8/10 & \textbf{10/10} \\
    Brush Box     & 2.5 & \textbf{9.5}  & 2/10 & 0/10 & 4/10 & \textbf{7/10} \\
    \bottomrule
  \end{tabular*}
  \vspace{-6pt}
\end{table}

\begin{figure}[t]
  \centering
  \includegraphics[width=\textwidth]{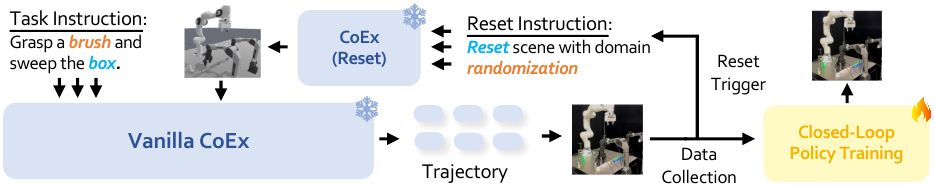}
  \vspace{-14pt}
  \caption{\textbf{Closed-loop data collection.} The agent solves the task (\emph{Vanilla \ours{}}) and its trajectory is recorded; after a reset, a reset instruction makes the same frozen agent re-issue the task from the new configuration (\emph{\ours{} (Reset)}), so the same program is reused across episodes. The recorded rollouts fine-tune a visuomotor policy.}
  \label{fig:flywheel}
  \vspace{-8pt}
\end{figure}

\noindent\textbf{A small distillation study.}
To test whether these rollouts are useful supervision, we collect them in a closed loop (Fig.~\ref{fig:flywheel}): once a program is verified, the frozen agent re-issues it from each new initial configuration, so a whole session reuses one round of exploration (scene resets were manual here; Appendix~\ref{sec:supp-distillation}). We fine-tune a $\pi_{0.5}$ policy~\citep{pi2025pi05} on 50 and on 200 \ours{} rollouts recorded on the real robot, against two references: 20 teleoperated demonstrations, and the standard co-training recipe of pairing those demonstrations with 200 rollouts of the \emph{same verified program} recorded in the twin rather than on the robot. Success grows steadily with the number of \ours{} rollouts, reaching 10/10 and 7/10 at 200 (Table~\ref{tab:data-collection}). Co-training on the simulated rollouts does not help here: at 3/10 and 0/10 it falls below the teleoperated demonstrations alone. With the program, the episode count and the evaluation protocol held fixed, the only thing separating that column from the last one is where the rollouts were recorded. The gap is consistent with what the policy consumes: $\pi_{0.5}$ acts from camera images, and rendered views differ from the real cameras in texture, lighting, and sensor noise, so simulated episodes train the visuomotor mapping on a distribution it never meets at test time. The program crosses that gap because it is written in terms of object poses; the pixels do not.

\section{Conclusion and Limitations}
\label{sec:conclusion}

\ours{}, hosted by the CoEnv framework, gives a coding agent what it needs to program real multi-arm systems: an executable digital twin of its own deployment workspace in which to write, run, and debug programs, grounded verification tools that tell it what went wrong, and a collision-checked path back to hardware. Real-robot experiments on five multi-arm tasks validate the pipeline against prior frameworks and against the same system without its code agent, and a small distillation study shows that the verified rollouts exploration leaves behind already suffice to fine-tune a visuomotor policy, and that recording them on the robot rather than in the twin is what makes them useful, so the agent's practice is not wasted. More broadly, the bottleneck is less the agent than the environment it can practice in: given a twin with inspectable state, an off-the-shelf code agent already programs heterogeneous multi-arm systems, and the data it leaves behind needs no sim-to-real bridge.

\noindent\textbf{Limitations.}\label{sec:limitations}
The twin covers rigid objects with pre-generated assets, fixed bases, and static scenes, and coordination is centralized, so we claim nothing about decentralized autonomy or scaling beyond three arms. Real-world execution is open-loop replay, so residual pose offsets remain a main source of failure. Our evaluation is modest in scale, and \ours{} and its ablation differ in backbone model as well as in loop structure.

\subsubsection*{Reproducibility Statement}
We describe the full system in \Secref{sec:method} and Appendix~\ref{sec:supp-implementation}: the perception models used for digital twin construction, the simulator, the foundation models, the primitive and verification interface exposed to the code agent (Fig.~\ref{fig:api}), and the hardware and control stack. The complete prompt structures of \ours{} and of the ablated baseline are reproduced in Appendix~\ref{sec:supp-prompts}, task and milestone definitions in Appendix~\ref{sec:supp-tasks}, the evaluation protocol in \Secref{sec:experimental-setup}, and the data collection and policy fine-tuning details of the distillation study in Appendix~\ref{sec:supp-distillation}. Per-task counts behind every figure are tabulated in Appendix~\ref{sec:supp-bars}. We will release our code, including the twin construction pipeline, the agent interface, and all prompts, upon publication. Because the agents rely on proprietary foundation models (GPT-5.6 Sol and Claude Code) whose behavior may change across versions, we will document the model versions used in our experiments together with the code.

\subsubsection*{Ethics Statement}
This work studies tabletop manipulation with fixed-base robot arms in a controlled laboratory setting. It involves no human-subject experiments and no personal or sensitive data. Letting a language-model agent write programs that move physical robots raises a safety concern, which our design addresses directly: every program is first executed and verified in a digital twin, and a trajectory is admitted to hardware only after a pre-execution swept-volume collision check. These mechanisms reduce but do not remove risk, so all real-robot experiments were run under human supervision with standard physical safeguards (workspace barriers and emergency stops), and we recommend the same for any deployment of this kind of system. We foresee no negative societal impact specific to this method beyond those common to robot automation in general.

\subsubsection*{AI Use Statement}
Foundation models are components of the method studied in this paper: GPT-5.6 Sol serves as the task planner, disambiguates object instances during digital twin construction, and drives the ablated baseline (\base{}), and Claude Code is the code agent inside \ours{} (\Secref{sec:method}). Their prompts are reproduced in Appendix~\ref{sec:supp-prompts}, and the rollouts used in the distillation study are, by design, generated by these agents. Beyond the method itself, we used an LLM-based assistant during manuscript preparation: for grammar checking and language polishing, including tightening and reorganizing the text and feedback on its presentation; for drafting prose that the authors then revised, including parts of the related-work discussion, the positioning of this work against prior real-to-sim paradigms, and several figure captions; for literature retrieval, that is, surfacing candidate related work and verifying every bibliography entry against its original source; for cross-checking the consistency of reported numbers between tables, figures, and prose; and for writing the plotting script of Fig.~\ref{fig:bars}. The research ideas, the system design and implementation, all experiments, and the analysis and interpretation of the results are the work of the authors. All AI-assisted text was reviewed and revised by the authors, who take full responsibility for the content of this paper; no AI system is listed as an author.

\bibliography{main}
\bibliographystyle{iclr2027_conference}

\newpage
\appendix

\section{Positioning with Respect to Prior Data Pipelines}
\label{sec:supp-positioning}

Table~\ref{tab:supp-positioning} expands Fig.~\ref{fig:motivation}. The column that matters is \emph{what a human must supply per task}: every prior pipeline needs either demonstrations or a reward, because the simulator can only replay or optimize behavior that something else has specified. \ours{} supplies that behavior itself, which is why the human input reduces to a task description, and why the simulator is never asked to produce a training image.

\begin{table}[h]
  \centering
  \caption{\textbf{Where the simulator sits in sim-assisted manipulation data pipelines.} ``Policy trains on'' refers to the data the deployed policy is fitted to. Rows: MimicGen~\citep{mandlekar2023mimicgen}, DexMimicGen~\citep{jiang2025dexmimicgen}, ComSim~\citep{qin2026comsim}, sim-and-real co-training~\citep{maddukuri2025simreal}, RialTo~\citep{torne2024rialto}, ACDC~\citep{dai2024automated}, HATS~\citep{lin2026hats}.}
  \label{tab:supp-positioning}
  \footnotesize
  \renewcommand{\arraystretch}{1.25}
  \setlength{\tabcolsep}{4pt}
  \begin{tabular*}{\textwidth}{@{\extracolsep{\fill}} l p{0.195\textwidth} p{0.235\textwidth} p{0.165\textwidth} c @{}}
    \toprule
    Method & Human input per task & What the simulator does & Policy trains on & Arms \\
    \midrule
    MimicGen    & $\sim$10 source demos      & replays and adapts them          & sim                      & 1--2 \\
    DexMimicGen & $\sim$5 source demos       & replays and adapts them          & sim                      & 2 \\
    ComSim      & 10 real demos              & classical sim + neural rendering & real + pseudo-real       & 2 \\
    Co-training & 10--30 real demos          & generates a large corpus         & sim + real (tuned ratio) & 1--2 \\
    RialTo      & demos + success condition  & environment for RL               & sim (+ randomization)    & 1 \\
    ACDC        & demos or reward            & generates digital cousins        & sim                      & 1 \\
    HATS        & an operator every episode  & not used                         & real                     & 4 \\
    \midrule
    \rowcolor{gray!6}
    \textbf{\ours{} (ours)} & \textbf{a task description} & \textbf{searches for and verifies a program} & \textbf{real} & \textbf{2--3} \\
    \bottomrule
  \end{tabular*}
\end{table}

\section{Task Descriptions}
\label{sec:supp-tasks}

\begin{figure}[h]
  \centering
  \includegraphics[width=\textwidth]{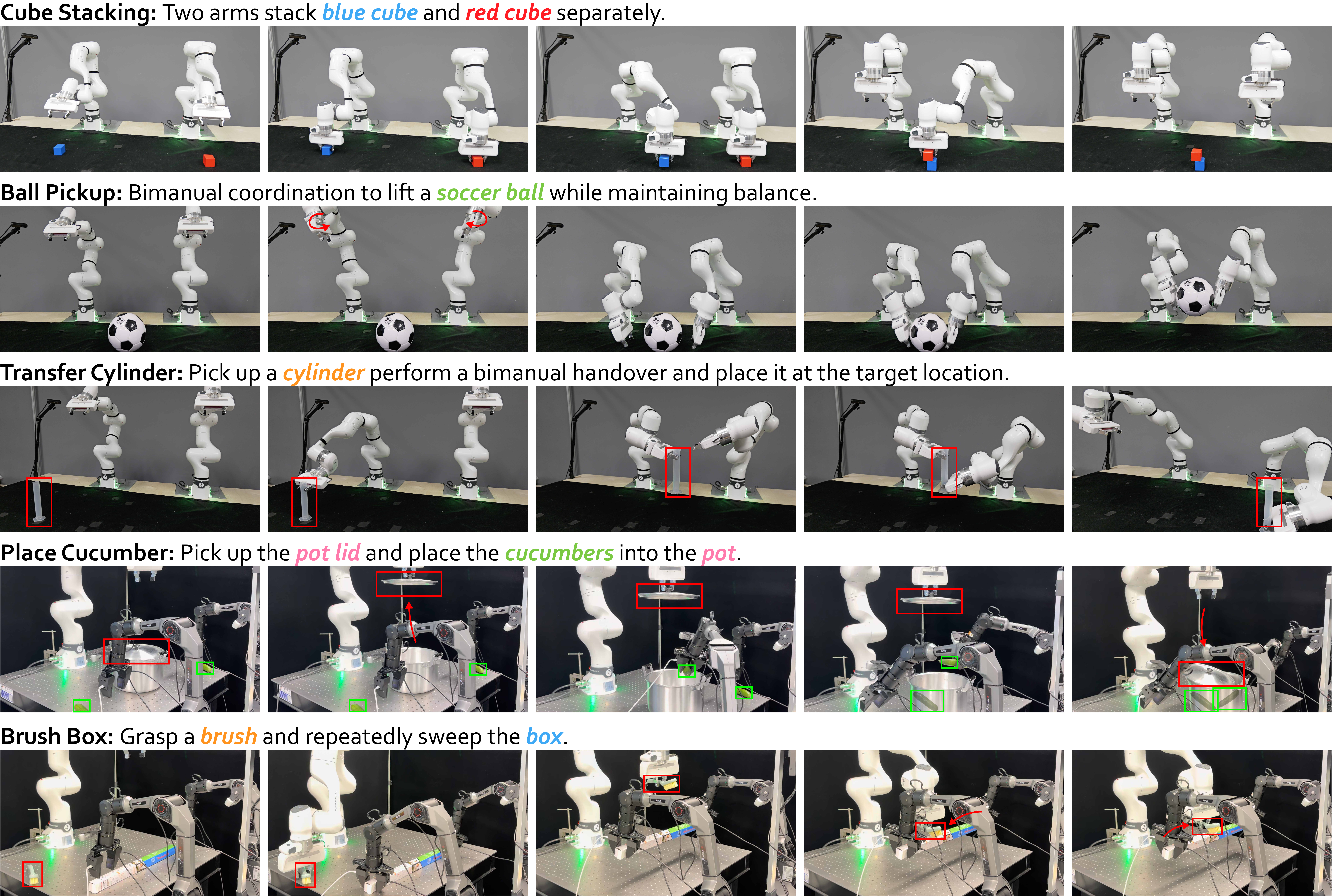}
  \caption{\textbf{Task suite.} Five multi-arm tasks with increasing coordination complexity. \textit{Top three rows:} two-arm tasks (Franka $\times$2): Cube Stacking, Ball Pickup, Transfer Cylinder. \textit{Bottom two rows:} three-arm tasks (Franka + AgileX Piper): Place Cucumber, Brush Box. Each row shows keyframes from a successful real-world execution.}
  \label{fig:tasks}
\end{figure}

Fig.~\ref{fig:tasks} shows keyframes from successful real-world executions of all five tasks. Table~\ref{tab:supp-task-desc} summarizes the five evaluation tasks and their objectives. Table~\ref{tab:supp-task-milestone} details the ordered subtask milestones ($S_i$) used for evaluation. Each milestone must be satisfied in order; the task succeeds only when the final milestone is achieved.

\begin{table}[h]
  \centering
  \vspace{-9pt}
  \caption{\textbf{Task descriptions.} Each task involves 2--3 robotic arms.}
  \label{tab:supp-task-desc}
  \footnotesize
  \renewcommand{\arraystretch}{1.15}
  \setlength{\tabcolsep}{4pt}
  \begin{tabular*}{\textwidth}{@{\extracolsep{\fill}} l @{\hspace{8pt}} p{0.80\textwidth} @{}}
    \toprule
    \textbf{Task} & \textbf{Description} \\
    \midrule
    \rowcolor{gray!6}
    \multicolumn{2}{l}{\textit{Two-Arm (Franka $\times$ 2)}} \\
    Cube Stacking     & Two arms stack a blue cube and a red cube separately. \\
    Ball Pickup       & Bimanual coordination to lift a soccer ball with balanced contact. \\
    Transfer Cylinder & Pick up a cylinder, perform a bimanual handover and place it at the target location. \\
    \midrule
    \rowcolor{gray!6}
    \multicolumn{2}{l}{\textit{Three-Arm (Franka + AgileX Piper $\times$ 2)}} \\
    Place Cucumber    & Pick up the pot lid and place the cucumbers into the pot. \\
    Brush Box         & Grasp a brush and repeatedly sweep the box. \\
    \bottomrule
  \end{tabular*}
  \vspace{-12pt}
\end{table}

\begin{table}[h]
  \centering
  \caption{\textbf{Subtask milestones and success criteria.} The last milestone of each task defines the overall success condition.}
  \label{tab:supp-task-milestone}
  \footnotesize
  \renewcommand{\arraystretch}{1.15}
  \setlength{\tabcolsep}{4pt}
  \begin{tabular}{lcp{0.58\textwidth}}
    \toprule
    \textbf{Task} & \textbf{Milestone} & \textbf{Criterion} \\
    \midrule
    \multirow{2}{*}{Cube Stacking}
      & $S_1$ & Both cubes are successfully grasped and lifted. \\
      & $S_2$ & The cubes are stably stacked. \\
    \midrule
    \multirow{2}{*}{Ball Pickup}
      & $S_1$ & Both arms reach the ball with contact poses. \\
      & $S_2$ & The ball is successfully lifted off the surface. \\
    \midrule
    \multirow{3}{*}{Transfer Cylinder}
      & $S_1$ & The cylinder is grasped and lifted by the first arm. \\
      & $S_2$ & The second arm successfully receives the cylinder. \\
      & $S_3$ & The cylinder is placed at the target location. \\
    \midrule
    \multirow{3}{*}{Place Cucumber}
      & $S_1$ & Pot lid is opened and held stably. \\
      & $S_2$ & Two cucumbers are picked up. \\
      & $S_3$ & Both cucumbers are placed inside the pot. \\
    \midrule
    \multirow{3}{*}{Brush Box}
      & $S_1$ & The brush is successfully grasped. \\
      & $S_2$ & The box is successfully grasped. \\
      & $S_3$ & The brush contacts the box in a sweeping motion. \\
    \bottomrule
  \end{tabular}
  \vspace{-12pt}
\end{table}

\section{Implementation Details}
\label{sec:supp-implementation}

Algorithm~\ref{alg:action-synthesis} in the main text formalizes \ours{}. The \base{} baseline shares Stage~I and replaces the program loop (lines 4--9) with a per-primitive \textit{execute $\to$ observe $\to$ verify $\to$ adapt} loop, described in \Secref{sec:supp-interactive}.

\subsection{Digital Twin Construction}
\label{sec:supp-twin}

\paragraph{3D asset generation.}
Rather than relying on noisy photogrammetric meshes, we generate simulation-ready meshes from a few reference images per object using Meshy Model\footnote{\url{https://www.meshy.ai/}}, scale-normalized to real dimensions, paired with low-polygon convex collision proxies, and assigned per-category physical attributes (mass, friction); authoring is one-shot per object class.

\paragraph{Multi-view object localization.}
For each view \(o_t^{(k)}\), Grounded SAM2~\citep{ren2024grounded} detects and segments target objects, and GPT-5.6 Sol~\citep{openai2026gpt56} disambiguates visually similar instances using spatial context (e.g., two same-shape cubes, or a brush partially occluded by a gripper). FoundationPose~\citep{wen2024foundationpose} then produces a 6-DoF pose \(p^{(k)} = (\mathbf{t}^{(k)},\mathbf{R}^{(k)})\) per view; we fuse poses across the \(K'\) detecting views \(\mathcal{K}'\) as
\begin{equation}
\hat{\mathbf{t}} = \tfrac{1}{K'} \sum_{k \in \mathcal{K}'} \mathbf{t}^{(k)}, \qquad
\hat{\mathbf{R}} = \mathcal{M}\!\bigl(\{\mathbf{R}^{(k)}\}_{k \in \mathcal{K}'}\bigr),
\label{eq:pose-fusion}
\end{equation}
arithmetically for translations and via quaternion averaging~\citep{markley2007averaging} on \(SO(3)\) for rotations, mitigating single-view occlusions and small extrinsic biases.

\paragraph{Iterative camera calibration.}
Exploiting the simulator's reset-ability, we refine camera extrinsics iteratively, rendering the current scene and comparing it to the real capture until reprojection error stabilizes at the millimeter range.

\subsection{Hardware and Robot Control}
\label{sec:supp-hardware}

\paragraph{Robot platforms.}
We use two robot platforms: the Franka Research~3 (7-DoF) and the AgileX Piper (6-DoF, dual-arm configuration). Both robots operate in joint position control mode. The Franka is controlled via Deoxys~\citep{zhu2022viola}, which provides a modular real-time control interface over the \texttt{libfranka} communication layer; we adopt its default interpolation scheme that generates smooth joint-space trajectories between waypoints. The AgileX Piper is controlled via the official Piper SDK~\citep{agilex2024pipersdk}; since the SDK provides only raw joint position commands, we implement linear interpolation between consecutive waypoints to ensure smooth and safe motion execution.

\paragraph{Perception setup.}
We deploy 2 calibrated Intel RealSense D435i cameras for the two-arm setting (Franka $\times$ 2) and 3 cameras for the three-arm setting (Franka + AgileX Piper $\times$ 2). All cameras are mounted at fixed positions around the workspace and provide synchronized RGBD streams for multi-view scene reconstruction (\Secref{sec:environment-digitization}).

\subsection{Agentic Exploration in the Digital Twin}

\subsubsection{Shared Infrastructure.}
\label{sec:supp-shared}

Both agents share a common action and perception layer built atop ManiSkill~\citep{tao2024maniskill3}.
We define four delta-based action primitives, namely \textsc{Move}, \textsc{Rotate}, \textsc{Grasp}, and \textsc{Release}, all specified relative to the current end-effector pose to eliminate the need for global coordinate calibration.
Each primitive accepts an arm index (\texttt{agent\_id} in the prompts) that can be a single integer or a list for synchronized multi-arm execution.
A state query API exposes the TCP pose, gripper aperture, and 6-DoF object poses, providing the numerical grounding for both agents.
A unified controller dispatches primitives to robot-specific IK solvers for Franka Research~3 (7-DoF) and AgileX Piper (6-DoF).
A virtual camera module renders $1920{\times}1080$ images from arbitrary viewpoints around the scene centroid.

\subsubsection{\base{}: Step-wise VLM Baseline.}
\label{sec:supp-interactive}

\base{} implements the closed-loop \textit{execute $\to$ observe $\to$ verify $\to$ adapt} cycle using GPT-5.6 Sol~\citep{openai2026gpt56} as the VLM. Execution proceeds in two phases.

\paragraph{Planning phase.}
The VLM performs multi-round visual analysis (up to $J_{\max}{=}6$ rounds), receiving a rendered scene image, structured state data, and a system prompt encoding shared manipulation knowledge (see Sec.~\ref{sec:supp-prompts}). It may request additional viewpoints via \texttt{CAMERA\_ORBIT} to observe the scene from multiple perspectives, implementing the adaptive multi-view aggregation of \Secref{sec:cognitive-planning-with-embodied-grounding}. The phase concludes with \texttt{PLANNING\_\allowbreak COMPLETE} and three structured outputs: \textit{key observations} (spatial findings carried forward as persistent context), \textit{checkpoints} (steps requiring visual verification with recommended camera angles), and an \textit{execution plan} (action sequence with robot assignments).

\paragraph{Execution phase.}
Given the plan $\mathcal{E} = \{e_1, \dots, e_L\}$, each step follows an \textit{observe--act--verify} loop: the VLM receives the current image and state, outputs structured reasoning followed by one or more actions. At checkpoint steps, the system enforces position, orientation, and visual verification before proceeding. Corrective actions are generated automatically when checks fail. To improve robustness, the system employs stuck pattern detection (injecting hints when repetitive low-magnitude actions are detected), post-action drift correction, and idle-arm stabilization for multi-arm tasks.

\subsubsection{\ours{}: Code-Driven Exploration.}
\label{sec:supp-iterative}

\ours{} implements the code generation and refinement loop of \Secref{sec:cognitive-planning-with-embodied-grounding}, using Claude Code~\citep{anthropic2025claudecode} as the code agent. Rather than issuing one action at a time, the agent generates a \emph{complete} Python program encoding the entire multi-arm execution logic.

\paragraph{System prompt and sandbox.}
The code agent receives a structured system prompt covering the task specification, API reference, manipulation knowledge base, collision avoidance rules, and a task-type-specific code template. Generated code runs in a sandboxed environment with a restricted namespace and timeout enforcement, ensuring reproducibility. A checkpoint function enables the code to capture multi-view images and state snapshots at critical execution points.

\paragraph{Iterative refinement.}
Each iteration $m$ resets the simulation to the same initial state and executes the generated program. Post-execution, the code agent analyzes execution logs and checkpoint images to identify root causes of failure (e.g., unreachable targets, failed grasps), then refines the code accordingly. This process repeats for up to $M_{\max}{=}10$ iterations. Each iteration is \emph{stateless}: the environment resets from scratch and the code must be self-contained.

\subsubsection{Comparison of the Two Agents.}

Table~\ref{tab:supp-mode-comparison} summarizes the key design differences. In the real-robot results (Table~\ref{tab:main_results}), \ours{} is ahead on all five tasks. What separates the two designs most clearly is cost and reusability (\Secref{sec:exp-distill}) and how success is retained across milestones (\Secref{sec:exp-real}).

\begin{table}[t]
  \centering
  \caption{\textbf{Design comparison of the \base{} baseline and \ours{}.}}
  \label{tab:supp-mode-comparison}
  \footnotesize
  \renewcommand{\arraystretch}{1.15}
  \setlength{\tabcolsep}{3pt}
  \begin{tabular*}{\textwidth}{@{\extracolsep{\fill}} lp{0.35\textwidth}p{0.35\textwidth} @{}}
    \toprule
    \textbf{Aspect} & \textbf{\base{}} & \textbf{\ours{} (ours)} \\
    \midrule
    Foundation model  & GPT-5.6 Sol (VLM)                   & Claude Code (code agent)          \\
    Decision granularity & One action per VLM call     & Full program per iteration        \\
    Feedback signal   & Real-time images + state       & Post-hoc checkpoints + state      \\
    API calls / task  & ${\sim}$30+ VLM calls          & ${\sim}$1--5 code generations      \\
    Error correction  & Adjust next action online      & Rewrite code offline              \\
    Planning          & Multi-round visual analysis    & Structured system prompt          \\
    Verification      & Checkpoint + visual + numerical& Checkpoint images + stdout logs   \\
    Advantage         & Reactive, adaptive coordination& Precise, coherent trajectories    \\
    \bottomrule
  \end{tabular*}
  \vspace{-8pt}
\end{table}

\subsection{Per-Task Counts behind Fig.~\ref{fig:bars}}
\label{sec:supp-bars}

Tables~\ref{tab:baseline_comparison} and~\ref{tab:ablation} give the per-task success counts (and, for the tool ablation, the subtask milestones) plotted in Fig.~\ref{fig:bars}.

\begin{table}[h]
  \centering
  \caption{\textbf{Comparison with prior VLM-planning frameworks} and our \base{} baseline on three representative tasks (easy two-arm, reactive two-arm handover, three-arm). SR ($x$/10); bold: best per task.}
  \label{tab:baseline_comparison}
  \footnotesize
  \renewcommand{\arraystretch}{1.2}
  \setlength{\tabcolsep}{6pt}
  \begin{tabular*}{0.92\textwidth}{@{\extracolsep{\fill}} l cccc @{}}
    \toprule
    Task & Code-as-Monitor & SimPACT & \base{} & \textbf{\ours{}} \\
    \midrule
    Cube Stacking      & 5/10 & 8/10 & 6/10 & \textbf{9/10} \\
    Transfer Cylinder  & 1/10 & 0/10 & 4/10 & \textbf{6/10} \\
    Brush Box          & 0/10 & 4/10 & 7/10 & \textbf{8/10} \\
    \midrule
    \rowcolor{gray!6}
    \textit{Average}   & 20\% & 40\% & 57\% & \textbf{77\%} \\
    \bottomrule
  \end{tabular*}
  \vspace{-6pt}
\end{table}

\begin{table}[h]
  \centering
  \caption{\textbf{Tool ablation (\base{}):} removing adaptive camera control or checkpoint verification.}
  \label{tab:ablation}
  \footnotesize
  \renewcommand{\arraystretch}{1.05}
  \setlength{\tabcolsep}{4pt}
  \begin{tabular*}{\textwidth}{@{\extracolsep{\fill}} l cccc cccc @{}}
    \toprule
    & \multicolumn{4}{c}{\textbf{w/o Adaptive Camera}} & \multicolumn{4}{c}{\textbf{w/o Checkpoint Verif.}} \\
    \cmidrule(lr){2-5} \cmidrule(lr){6-9}
    Task & $S_1$ & $S_2$ & $S_3$ & SR & $S_1$ & $S_2$ & $S_3$ & SR \\
    \midrule
    Cube Stacking      & 6/10 & 5/10 & {--} & 5/10 & 4/10 & 4/10 & {--} & 4/10 \\
    Ball Pickup        & 10/10 & 6/10 & {--} & 6/10 & 10/10 & 2/10 & {--} & 2/10 \\
    Transfer Cylinder  & 6/10 & 2/10 & 0/10    & 0/10 &  6/10 &  0/10 &  0/10    &  0/10 \\
    Place Cucumber     & 9/10 & 6/10 & 4/10    & 4/10 &  6/10 & 6/10 & 4/10    & 4/10 \\
    Brush Box          & 10/10 & 8/10 & 0/10    & 0/10 &  9/10 &  8/10 &  0/10    & 0/10 \\
    \midrule
    \rowcolor{gray!6}
    \textit{Avg.}  & {82\%} & {54\%} & {13\%} & {30\%} & {70\%} & {40\%} & {13\%} & 20\% \\
    \rowcolor{blue!6}
    \textit{Full tools}  & {88\%} & {60\%} & {50\%} & \textbf{50\%} &  {88\%} & {60\%} & {50\%} & \textbf{50\%} \\
    \bottomrule
  \end{tabular*}
  \vspace{-6pt}
\end{table}

\section{Experience Reuse: Distillation Study Details}
\label{sec:supp-distillation}

Fig.~\ref{fig:flywheel} in the main text illustrates how \ours{} turns exploration into data: a program verified in the twin is executed on the robot, the scene is reset between episodes, and the recorded rollouts are stored for policy training. This section gives the protocol of the study in \Secref{sec:exp-distill}.

\subsection{Demonstration Collection Protocol}
For each of the two tasks (\textit{Cube Stacking} and \textit{Brush Box}) we collected 200 episodes on the real robot with \ours{}. An episode enters the dataset only if the task is completed; the 50-rollout dataset is a subset of the 200-rollout one. In this study the initial configurations were chosen by hand and the scene was reset manually between episodes, to make sure every episode started from a valid, known configuration; the reset instruction of Fig.~\ref{fig:flywheel} then re-issues the task from that configuration. Each episode records three RGB camera streams at $640{\times}480$ and $30$~Hz together with the joint positions of all arms; actions are joint-angle targets. The 20 reference demonstrations were collected by human teleoperation with GELLO~\citep{wu2024gello}, a low-cost leader-arm interface, and recorded in the same format.

\subsection{Policy Fine-Tuning and Evaluation}
We start from the publicly released pre-trained $\pi_{0.5}$ weights~\citep{pi2025pi05} and fine-tune one policy per task and per dataset (20 teleoperated, 50 \ours{}, 200 \ours{}), for 30k steps with batch size 64 on 8 H100 GPUs; all other settings follow the released fine-tuning recipe, and no data augmentation is used. The policy observes the same three camera views and joint positions as recorded during collection, and outputs joint-angle targets for all arms. Every policy is evaluated for 10 real-robot trials on its task, with the same evaluation protocol as in \Secref{sec:experimental-setup}.

\subsection{Experience Storage and Reuse}
The framework keeps the outcome of every verified episode. Rather than storing raw programs, the agent itself summarizes what it learned (strategies and grasp parameters that survived verification, and the causes of failed rounds) into short notes, which are stored and provided to \ours{} through the \emph{Operational Hints} field of its prompt (\Secref{sec:supp-code-prompts}) in later runs. We do not ablate this mechanism separately.

\section{Failure Analysis}
\label{sec:supp-failure}

We observe three recurring failure modes in the real-robot experiments; Fig.~\ref{fig:supp-failure} shows representative cases. We describe them qualitatively and note a mitigation for each.

\subsection{Sim-to-real positional discrepancy}
\begin{figure}[t]
  \centering
  \includegraphics[width=\textwidth]{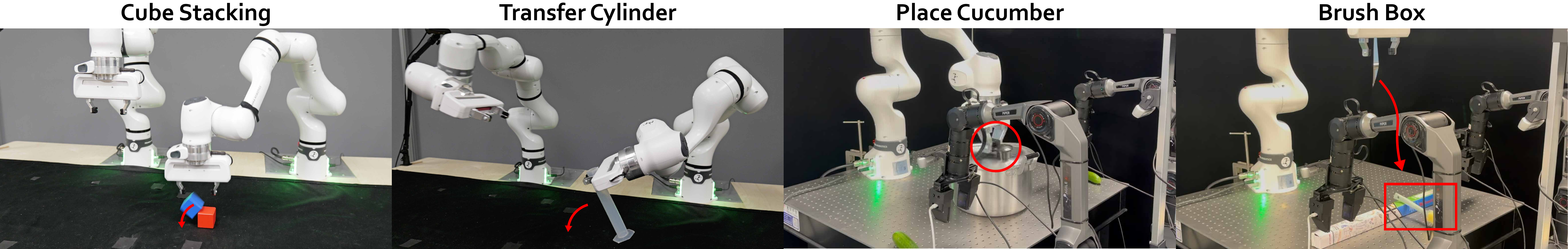}
  \caption{\textbf{Representative failure cases.} From left to right: (1)~\textit{Cube Stacking}: the stacked cube slips off due to sim-to-real positional offset at placement; (2)~\textit{Transfer Cylinder}: the receiving arm fails to align with the handing arm's gripper during handover, causing the cylinder to drop; (3)~\textit{Place Cucumber}: the pot lid is grasped at an unstable point (red circle), causing it to fall during the holding phase; (4)~\textit{Brush Box}: planning error causes the brush to be dropped (red rectangle) during moving phase.}
  \label{fig:supp-failure}
\end{figure}

This mode affects all five tasks to varying degrees. The root cause is residual error in camera extrinsic calibration and object pose estimation, which introduces millimeter-level spatial offsets between the planned simulation trajectory and the actual real-world execution. In \textit{Cube Stacking} (Fig.~\ref{fig:supp-failure}, left), the offset manifests at the final placement step: the arm positions the cube slightly off-center relative to the base cube, causing it to slide off after release. In \textit{Ball Pickup}, both arms must converge on the ball simultaneously, and even small per-arm offsets (${\sim}$2--3\,mm) compound into a gap that prevents stable bimanual contact. In \textit{Place Cucumber}, the narrow pot opening (${\sim}$12\,cm diameter) leaves minimal tolerance for insertion error, causing cucumbers to collide with the rim rather than entering cleanly. A lightweight residual correction at contact, or online re-calibration of camera extrinsics from fiducials between trials, would absorb these millimeter-level errors before they propagate to contact failures.

\subsection{Planning loop stagnation}
\label{sec:supp-stagnation}
In the step-wise agent, the VLM planner sometimes fixates on a specific grasp strategy that has already failed, repeatedly attempting minor pose adjustments without exploring fundamentally different approaches (e.g., switching from a top-down grasp to a side grasp). Our stuck pattern detection mechanism (Sec.~\ref{sec:supp-interactive}) mitigates but does not fully eliminate this issue. In code-driven exploration, the code agent occasionally produces only incremental parameter changes between iterations (e.g., shifting a target position by 1\,mm) without addressing the underlying geometric constraint violation, exhausting the iteration budget $M_{\max}$ without convergence. Diversity-promoting prompts after repeated failures, or an explicit failure memory that steers the agent away from strategies that have already failed, are natural remedies.

\subsection{Agent-specific limitations}
In the step-wise agent, cumulative drift is the primary concern for long action sequences. Each VLM-issued action introduces a small tracking error from the PD controller, and over 20+ sequential actions, these errors accumulate to produce noticeable end-effector drift. This is especially problematic in \textit{Brush Box} (Fig.~\ref{fig:supp-failure}, right), where the sweeping motion requires sustained contact between the brush and box over multiple strokes. By the third or fourth stroke, the brush may have drifted several millimeters away from the box surface. For \ours{}, the demanding cases are those in which one arm's motion depends on where another arm actually ended up, as in the \textit{Transfer Cylinder} handover (Fig.~\ref{fig:supp-failure}, second from left): the program must anticipate the hand-over pose instead of adapting to it online. Letting a program hand such reactive sub-steps to step-wise control, while keeping the rest at the program level, is a promising hybrid.

\section{Prompt Design}
\label{sec:supp-prompts}

This section presents the prompt designs for both agents. We first describe the shared domain knowledge embedded in both prompts (\S\ref{sec:supp-shared-knowledge}), then detail the prompt structures of the step-wise baseline (\S\ref{sec:supp-vlm-prompts}) and of code-driven exploration (\S\ref{sec:supp-code-prompts}).

\subsection{Shared Domain Knowledge}
\label{sec:supp-shared-knowledge}

Both agents embed a common manipulation knowledge base that provides the foundation model with the minimal physical specifications needed to operate the heterogeneous robot fleet. Table~\ref{tab:supp-domain-knowledge} summarizes the categories of domain knowledge and their roles. Importantly, this knowledge base contains only \emph{generic robot specifications and physics constraints} (e.g., gripper dimensions, workspace limits, coordinate semantics); it does not include any task-specific solutions or pre-computed trajectories. The model must still reason about object geometry, plan grasp strategies, and coordinate multi-arm actions autonomously based on real-time observations.

\begin{table}[t]
  \centering
  \caption{\textbf{Shared domain knowledge} embedded in the system prompts of both agents. All entries are generic robot/physics specifications, not task-specific solutions.}
  \label{tab:supp-domain-knowledge}
  \footnotesize
  \renewcommand{\arraystretch}{1.15}
  \setlength{\tabcolsep}{4pt}
  \begin{tabular}{lp{0.65\textwidth}}
    \toprule
    \textbf{Category} & \textbf{Content} \\
    \midrule
    Gripper specifications & Maximum opening width and finger length for each robot type (Franka, Piper). \\
    Orientation semantics & Delta-based rotation convention; mapping from pitch/yaw values to physical gripper directions. \\
    Grasp strategy guidelines & Shape-conditioned heuristics (top-down for flat objects, horizontal for upright cylinders, cooperative bimanual for oversized objects); edge-grasp safety margins; fingertip positioning rules (overshoot for horizontal grasps, low-$z$ target for flat objects). \\
    Workspace limits & Per-robot reachable workspace ranges. \\
    Collision avoidance & Awareness that arm links sweep a volume beyond the TCP; minimum clearance thresholds; retreat-toward-base principle for clearing shared workspace. \\
    \bottomrule
  \end{tabular}
\end{table}

\subsection{\base{} Baseline: VLM Prompts}
\label{sec:supp-vlm-prompts}

The step-wise baseline uses two distinct prompts sent to GPT-5.6 Sol with multi-modal (text + image) input: a \emph{planning phase prompt} for multi-round scene analysis, and an \emph{execution phase prompt} for closed-loop action generation. We present each prompt's overall structure (with short sections inlined), then expand the output format blocks separately.

\subsubsection{Planning Phase Prompt.}

The following box shows the full planning prompt skeleton. Sections referencing Table~\ref{tab:supp-domain-knowledge} provide the shared domain knowledge described above; the output format is given in Prompt~2.

\begin{promptbox}[Planning Phase: Full Prompt Structure]
You are a robotics expert planning a manipulation task.
Analyze the scene and create a HIGH-LEVEL strategy.

# Task
{task_name}: {task_description}
Success criteria: {success_criteria}
# Operational Considerations
- {hint_1}   (task-specific tips, e.g., reach limits)
- {hint_2}

# Current State
  Robot {name} (id={id}, type={type}):
    Position: (x, y, z)
    Euler (RPY): (r, p, y)
    Gripper: {open/closed}
  Object {name} ({attr_name}):
    Position: (x, y, z)
    Size: {sx} x {sy} x {sz} m

# Visual Observation Strategy
You MUST actively observe the scene from multiple angles
before declaring planning complete:
- Request at least 2 different camera angles
- Use top-down view (pitch=1.2+) for XY layout and
  object orientations
- Use side views (yaw=0, +/-1.57, +/-3.14) for heights
  and gripper clearances
- Cross-reference visual observations with numerical
  state data (positions, sizes)
- If visual contradicts numerical data, trust the visual

# Per-Object Scene Analysis
Before planning any grasp, analyze EACH object independently:
determine its shape, long-axis direction, and the graspable
width at the planned grasp point. Set gripper yaw so fingers
close along the narrowest dimension.

# Gripper & Grasp Reference
  [Domain knowledge: see (*@Table~\ref{tab:supp-domain-knowledge}@*)]

# Multi-Arm Collision Avoidance
  [Domain knowledge: see (*@Table~\ref{tab:supp-domain-knowledge}@*)]

# Output Format
  => See Prompt 2
\end{promptbox}

\begin{promptbox}[Planning Output Format]
<thinking>
Your analysis:
- What you see in the image

Per-Object Orientation Analysis:
| Object | Shape | Long Axis | Grasp Point Location | Graspable Dim at Grasp Point | Chosen Yaw | Reason |

For each object, explain:
- Why this yaw was chosen for THIS specific object
- What dimension the fingers will close along
- Grasp strategy for each object
- Pre-grasp verification strategy
- Potential collision risks and mitigations
</thinking>

<next_action>
Either:
1. Request a different view (RECOMMENDED):
   {"type": "CAMERA_ORBIT",
    "params": {"yaw": X.XX, "pitch": X.XX},
    "reason": "why this angle helps"}

2. Or declare planning complete:
   PLANNING_COMPLETE

   <key_observations>
   Critical findings for execution:
   - Object positions and orientations
   - Chosen grasp strategy
   - Key constraints (clearances, collision avoidance)
   </key_observations>

   <checkpoints>
   Steps requiring verification before proceeding:
   - CP1: what to verify
     Position: xyz within 0.02m?
     Orientation: pitch/yaw within 0.1 rad?
     Visual: camera confirms object between fingers?
     Recommended view: yaw=X.XX, pitch=X.XX
   </checkpoints>

   <execution_plan>
   Numbered steps with approximate targets:
   1. [Robot X] MOVE + ROTATE - reason
   2. [Robot X] GRASP - CHECKPOINT CP1
   3. [Robot X] MOVE (lift) + MOVE (transport) + RELEASE

   Multi-robot coordination:
   MERGE synchronized actions into ONE step
   SEPARATE for different actions or verification
   </execution_plan>
</next_action>
\end{promptbox}

\subsubsection{Execution Phase Prompt.}

The following box shows the full execution prompt skeleton. The output format is given in Prompt~4.

\begin{promptbox}[Execution Phase: Full Prompt Structure]
You are executing a manipulation task. Your job is to OBSERVE
the scene and CALCULATE the exact movements needed.

# Task
{task_name}: {task_description}

# Current State
  [Same format as planning, plus inter-agent distances]

# Strategy from Planning
{key_observations from planning phase}
NOTE: Delta values from planning may be STALE -- always
recompute from CURRENT state data.

# Execution Plan (Step {current}/{total})
  [check] Step 1: [Franka] MOVE to lid (done)
  [->]    Step 2: [Franka] GRASP - CHECKPOINT (current)
  [o]     Step 3: [Franka] MOVE lift lid (pending)
  Current Goal: {current_step_description}

# Actions
Camera:
- CAMERA_ORBIT(yaw, pitch)
Robot (always include agent_id):
- MOVE(delta_x, delta_y, delta_z, agent_id)
- ROTATE(delta_yaw, agent_id, delta_pitch=0, delta_roll=0)
- GRASP(target_width, agent_id): -0.6 Franka, -0.8 Piper
- RELEASE(agent_id)
  agent_id: single int OR list for synchronized actions

# Gripper Geometry & Workspace Limits
  [Domain knowledge: see (*@Table~\ref{tab:supp-domain-knowledge}@*)]

# Pre-GRASP Verification
Before GRASP, verify numerically AND visually:
1. Position: error > 0.02m -> MOVE to correct first
2. Orientation: error > 0.1 rad -> ROTATE first
3. Visual: CAMERA_ORBIT to confirm object between fingers

# Multiple Actions in One Output
Combine when safe (e.g., ROTATE -> MOVE -> GRASP).
Do NOT combine when visual verification is needed.

# Output Format
  => See Prompt 4
\end{promptbox}

\begin{promptbox}[Execution Output Format]
<observation>
What you see:
- Where are the target object and gripper?
- Cross-check: does the image match numerical state data?
- Any collision risks?
</observation>

<reasoning>
1. Goal of current step
2. Read orientation from state data: quote the EXACT current
   yaw and pitch. Compare with target.
   If already within 0.1 rad -> do NOT rotate.
3. Position: current vs target -> required delta_x/y/z
4. Orientation: if drift > 0.1 rad -> corrective ROTATE
5. If about to GRASP: position < 0.02m? orientation < 0.1
   rad? visually verified? If any fails -> fix first
6. Decision and action
</reasoning>

<action>
Single action:
{"type": "ACTION_TYPE", "params": {...}, "step_done": false}

OR multiple sequential actions:
[
  {"type": "ROTATE", "params": {...}, "step_done": false},
  {"type": "MOVE", "params": {...}, "step_done": false},
  {"type": "GRASP", "params": {...}, "step_done": true}
]
</action>
\end{promptbox}

\subsection{\ours{}: Code Agent Prompt}
\label{sec:supp-code-prompts}

Code-driven exploration uses a single comprehensive system prompt. The following box shows the full prompt skeleton; sections marked with ``$\Rightarrow$ \textit{Prompt~X}'' are expanded in the corresponding prompts below.

\begin{promptbox}[Code Agent: Full Prompt Structure]
You are a robot manipulation expert. Your job is to write and
iteratively refine Python code that controls multi-arm robots
in a physics simulation.

# TASK INFORMATION
Task: {task_name}
Description: {task_description}
Success Criteria: {success_criteria}
Robots:
  ID {id}: {type} ({n}-DOF), Name: "{name}", Role: {role}
Objects:
  {attr_name}: "{name}", Size: {x}x{y}x{z}m, {description}
Operational Hints:
  - {hint_1}   (task-specific tips from prior experience)

# API REFERENCE
  => See Prompt 6

# Coordinate System
X: +right/-left | Y: +away/-toward | Z: +up/-down
Table surface: z ~ 0

# Manipulation Knowledge
  [Domain knowledge: see (*@Table~\ref{tab:supp-domain-knowledge}@*)]

# Robot Workspace & Reach Limits
- Franka: reach ~0.855m. At XY > 0.5m, Z severely limited.
- Piper: reach ~0.3m. Cannot raise above initial height.
- CRITICAL: MOVE to unreachable target silently stops at
  closest reachable point. MUST print diagnostics to detect.

# Diagnostic Printing
After critical MOVE -- print target vs actual position:
  state = robot.get_robot_state(0)
  print(f"TARGET: {t}, ACTUAL: {state['position']},
         GAP: {np.linalg.norm(...):.4f}")
  Large gap = arm hit joint limits.
After GRASP + LIFT -- print object z before/after:
  No z change = grasp missed.

# CODE RULES
1. ALWAYS read current state before computing deltas.
2. Call checkpoint() at critical moments.
3. Code must work for ANY initial state -- no hardcoding.
4. Use numpy (np) for vector math.
5. Each round resets from scratch -- code is self-contained.

# CODE TEMPLATE
  => See Prompt 7

# WORKFLOW (6 steps)
  => See Prompt 8
\end{promptbox}

\begin{promptbox}[API Reference]
## State Queries
robot.get_robot_state(robot_id: int) -> dict
  Returns: {"robot_id": int, "position": [x,y,z],
            "orientation": [qw,qx,qy,qz],
            "euler_rpy": [roll,pitch,yaw], "gripper": float}
  position: TCP fingertip in world coordinates (meters)
  gripper: 1.0 = open, -1.0 = fully closed
robot.get_all_robot_states() -> Dict[int, dict]
robot.get_object_state(attr_name: str) -> dict
  Returns: {"name", "position", "orientation"}
robot.get_all_object_states() -> Dict[str, dict]

## task_config Fields
task_config.robots  -- list of RobotConfig:
  .robot_id, .robot_type ("franka"/"piper"), .name, .role
task_config.objects -- list of ObjectConfig:
  .attr_name, .name, .size_x, .size_y, .size_z, .description

## Action Execution
robot.execute_action(action: dict) -> bool
  MOVE:    {"type":"MOVE", "params":{"delta_x":f, "delta_y":f,
            "delta_z":f, "agent_id":int|list}}
  ROTATE:  {"type":"ROTATE", "params":{"delta_yaw":f,
            "delta_pitch":f, "delta_roll":f,
            "agent_id":int|list}}
  GRASP:   {"type":"GRASP", "params":{"target_width":f,
            "agent_id":int|list}} // -0.6 Franka, -0.8 Piper
  RELEASE: {"type":"RELEASE", "params":{"agent_id":int|list}}
  Multi-robot: agent_id as list for synchronized actions.
  Params can be scalars (broadcast) or lists (per-robot).

## Checkpoint
checkpoint(name, checkpoint_type="generic", notes="",
           suggested_views=None)
  Types: "grasp", "lift", "handover", "place", "generic"
  Auto-generated: final_state, EXECUTION_ERROR, TIMEOUT_ERROR
\end{promptbox}

\begin{promptbox}[Code Template (2 Piper + 1 Franka)]
# Available: robot, task_config, checkpoint, np, time, print

# 1. Read current state
r0 = robot.get_robot_state(0)  # Piper-0
r1 = robot.get_robot_state(1)  # Piper-1
r2 = robot.get_robot_state(2)  # Franka
objects = robot.get_all_object_states()

# 2. Open all grippers
robot.execute_action(
  {"type": "RELEASE", "params": {"agent_id": 0}})
robot.execute_action(
  {"type": "RELEASE", "params": {"agent_id": 1}})
robot.execute_action(
  {"type": "RELEASE", "params": {"agent_id": 2}})

# 3. Move to target (compute deltas from current position)
# For Piper: GRASP target_width = -0.8
# For Franka: GRASP target_width = -0.6

# 4. Grasp, lift, checkpoint ...
# checkpoint("step_name", checkpoint_type="generic",
#            notes="Describe what to verify here")
\end{promptbox}

\begin{promptbox}[Iterative Workflow]
Step 1: Understand the API
  The API REFERENCE above is the complete guide.

Step 2: Make a Plan
  Save to {workspace}/plan.md:
  - Grasp strategy per object (approach, orientation, yaw)
  - Action sequence and multi-arm coordination
  - Collision risks
  Optional: run exploration scripts (don't count as rounds).

Step 3: Write Code and Run
  Write to {code_file}. Code runs to completion in one shot
  -- NO real-time visual feedback. checkpoint() captures
  images for post-run review only.

Step 4: Analyse Results
  Output directory contains:
  - summary.json: error status, final states, checkpoints
  - execution_stdout.txt: all print() output
  - initial_state/: front.png, top.png, side.png
  - checkpoints/<name>/: images + states.json

  Analysis checklist (EVERY round):
  1. Read stdout -> target vs actual position gaps
  2. Read summary.json -> did objects actually move?
  3. View key checkpoint images (pre_grasp, post_lift)
  4. Identify ROOT CAUSE, not symptoms

Step 5: Iterate
  Fix code, run next round. Environment resets each time.

Step 6: Final Report -> {workspace}/final_report.json
\end{promptbox}

\end{document}